\documentclass[runningheads]{llncs}

\usepackage{iciap}

\usepackage{iciapabbrv}

\usepackage{graphicx}
\usepackage{booktabs}
\usepackage[table,dvipsnames]{xcolor}
\usepackage{multirow}
\usepackage{color,colortbl,tabularx}
\usepackage[accsupp]{axessibility}  

\usepackage{hyperref}

\usepackage{xcolor} 
\usepackage{soul} 
\usepackage{orcidlink}

\usepackage{makecell}
\usepackage{multirow}
\usepackage{hhline}

\def\geneExpression{\mathbf{x_g}}
\newcommand{\geneExpressionOut}[1]{\mathbf{x_g^{#1}}}
\def\wholeSlideImage{\mathbf{I}}
\def\patch{\mathbf{x_p}}
\def\textualDescription{\mathbf{x_t}}
\def\patchEmbeddings{\mathbf{E_{img}}}
\def\modulatedPatchEmbeddings{\mathbf{E_{img}^{FiLM}}}
\def\transformerPatchEmbeddings{\mathbf{E_{img}^\prime}}
\def\textTokenEmbeddings{\mathbf{E_{text}}}
\def\patchCLS{\mathbf{e_{img}^{CLS}}}
\def\trasformerPatchCLS{\mathbf{e_{img}^{CLS^\prime}}}
\def\textCLS{\mathbf{e_{text}^{CLS}}}
\newcommand{\multimodalEmbedding}[1]{\mathbf{\hat{e}^{#1}}}
\def\noise{\mathbf{z}}
\newcommand{\normalDistribution}[1]{\mathcal{N}(0, I_{#1})}
\def\textSpace{\mathcal{T}}
\def\patchSize{P}
\def\numPatches{N}
\def\numTokens{M}
\def\embeddingDim{d}
\def\methodName{GeMM-GAN}

\begin{document}

\title{\methodName{}: A Multimodal Generative Model Conditioned on Histopathology Images and Clinical Descriptions for Gene Expression Profile Generation.} 

\titlerunning{\methodName{}: A Multimodal Generative Model for Gene Expression Profile}
\author{Francesca Pia Panaccione\orcidlink{0009-0005-8007-963X} \and
Carlo Sgaravatti\orcidlink{0009-0001-4962-5365} \and
Pietro Pinoli\orcidlink{0000-0001-9786-2851}}

\authorrunning{F.~Panaccione et al.}

\institute{DEIB - Dipartimento Elettronica, Informazione e Bioingegneria, Politecnico di Milano, Milan, Italy \\
\email{francescapia.panaccione@polimi.it} \email{carlo.sgaravatti@polimi.it} \\
\email{pietro.pinoli@polimi.it}}

\maketitle

\begin{abstract}

Biomedical research increasingly relies on integrating diverse data modalities, including gene expression profiles, medical images, and clinical metadata. While medical images and clinical metadata are routinely collected in clinical practice, gene expression data presents unique challenges for widespread research use, mainly due to stringent privacy regulations and costly laboratory experiments.
To address these limitations, we present \methodName{}, a novel Generative Adversarial Network conditioned on histopathology tissue slides and clinical metadata, designed to synthesize realistic gene expression profiles. \methodName{} combines a Transformer Encoder for image patches with a final Cross Attention mechanism between patches and text tokens, producing a conditioning vector to guide a generative model in generating biologically coherent gene expression profiles. We evaluate our approach on the TCGA dataset and demonstrate that our framework outperforms standard generative models and generates more realistic and functionally meaningful gene expression profiles, improving by more than 11\% the accuracy on downstream disease type prediction compared to current state-of-the-art generative models. Code will be available at: \url{https://github.com/francescapia/GeMM-GAN} 

\end{abstract}

\section{Introduction}
\label{sec:intro}


The integration of heterogeneous biomedical data, ranging from clinical records and histopathology images to gene expression profiles, has the potential to transform our understanding of disease mechanisms and enable more personalized healthcare. Yet, in practical settings, the availability of these data modalities is highly uneven. Clinical metadata and histopathology slides are routinely collected across institutions, whereas transcriptomic profiles remain limited due to their cost, privacy implications, and lack of standardization~\cite{liu2024contrastive}. Most medical centers can readily collect tissue slides and corresponding whole slide images (WSIs), but may lack the infrastructure and expertise to perform RNA sequencing experiments for gene expression analysis. However, gene expression analysis is often essential for guiding personalized treatment selection. For example, PAM50-guided treatment decisions in breast cancer have led to significant improvements in patient survival outcomes by enabling better risk stratification and optimal therapy selection for each tumor subtype \cite{ohnstad2017prognostic}. 

Deep generative models offer a promising solution for addressing this data gap. Recent progress in this domain has demonstrated their ability to model complex, high-dimensional biological data and produce realistic, privacy-preserving synthetic gene expression profiles\cite{van2024synthetic}. However, prior approaches to transcriptomic generation have typically relied on simplified inputs (either low-dimensional vectors or pathology images alone) failing to fully capture the rich context available in real-world clinical settings \cite{vinas2022adversarial,lacan2023gan}. Other approaches \cite{zheng2024digital,schmauch2020deep}, instead, model the problem as a prediction task, thus predicting the gene expression profile from histopathology images, without any generative modelling, missing the opportunity to leverage cross-modal relationships for synthetic data generation.

To overcome these limitations, we introduce  \textbf{\methodName{}}, a Multimodal Generative Adversarial Network conditioned both on histopatology images and clinical descriptions of the patients to generate biologically coherent gene expression profiles \emph{in silico}. To the best of our knowledge, we are the first addressing the problem of conditioning a generative model of gene expressions on both images and text. Our method builds on prior work on gene expression prediction from Whole Slide Images (WSIs), leveraging patch-based preprocessing via Otsu thresholding~\cite{schmauch2020deep} and Transformer-based modelling of image embeddings~\cite{zheng2024digital}. We furthermore study how to integrate the textual modality into this architecture, by leveraging Feature-wise Linear Modulation (FiLM) \cite{perez2018film} of the patch embeddings and by desinging a Cross-Attention mechanism to extract meaningful information from both modalities to effectively condition our generative model. We then leverage a Wasserstein Generaive Adversarial Network with Gradient Penalty (WGAN-GP) \cite{gulrajani2017} to generate gene expression profiles. 

We evaluate \methodName{} using standard metrics from the generative modelling literature, such as distributional alignment and downstream predictive performance, to demonstrate its ability to produce realistic, functionally meaningful gene expression profiles. 

Our contribution can be summarized as:
\begin{itemize}
    \item We introduce a novel multimodal framework for the generation of gene expression profiles conditioned jointly on histopathology images and clinical descriptions.
    \item We explore cross-modal fusion strategies (FiLM and cross-attention) for integrating textual and visual information into a shared latent representation.
    \item We show that our model can generate gene expression profiles that preserve biologically meaningful patterns and can support downstream predictive tasks, improving current state-of-the-art performance.
\end{itemize}

\section{Related Work}

Recent efforts at the intersection of computational pathology and transcriptomics have focused primarily on the prediction of gene expression values from histopathology images using deep learning. For example, Schmauch et al. \cite{schmauch2020deep} and Wang et al. \cite{wang2025deep} developed models to predict RNA-Seq profiles directly from WSIs. Similarly, Chlis et al. \cite{chlis2020predicting} predicted the expression profile of every cell in an imaging flow cytometry experiment. Although effective, these approaches are inherently discriminative, aiming to map visual features to molecular outputs. They do not address the underlying limitations of data scarcity and privacy.

Parallel research streams have explored the development of generative models for transcriptomics. For instance, Vinas et al. \cite{vinas2022adversarial}, Lacan et al. \cite{lacan2023gan}, and Panaccione et al. \cite{panaccione2025biogan} evaluated GANs for generating synthetic omics data. Yet, these approaches operate within the molecular domain, without incorporating imaging or clinical modalities. Conversely, image generation from transcriptomics has been investigated by Tariq et al \cite{carrillo2023rna}. using diffusion models. 
However, this approach does not address the challenge of augmenting transcriptomic datasets, and the multimodal framework remains less comprehensive compared to ours.

To the best of our knowledge, no existing approach generates gene expression profiles conditioned simultaneously on histopathology images and structured clinical metadata. Our framework, \methodName, is the first to integrate these heterogeneous modalities as conditioning signals in a generative model for transcriptomic synthesis.

\section{Problem Formulation}

In this work, we address the problem of generating biologically plausible gene expression profiles conditioned on multimodal inputs comprising Whole Slide Images (WSIs) and clinical descriptions of the patient. Specifically, let $\geneExpression \in \mathbb{R}^g$ be a gene expression profile with $g$ genes, $\wholeSlideImage \in \mathbb{R}^{W \times H \times 3}$ a WSI, where $W$ and $H$ are the sizes of the slide and $\textualDescription \in \textSpace$ the textual description. As a WSI is usually characterized by very high dimensions for $W$ and $H$, we denote with $\patch \in \mathbb{R}^{\patchSize \times \patchSize \times 3}$ a single patch of the WSI, and thus represent the WSI as a set of patches $\{\patch^{(i)}\}$. Our goal is to learn a conditional generative model $G(\geneExpression | \{\patch^{(i)}\}, \textualDescription)$ that captures the joint distribution of gene expression patterns given the visual and textual context. 
More in detail, we aim to learn a model that enables:
\begin{itemize}
    \item Conditional generation: synthesizing gene expression profiles consistent with provided image and text inputs.
    \item Multimodal conditioning: integrating visual features and textual semantics to inform gene-level patterns.
    \item Biological fidelity: producing outputs that reflect realistic and biologically coherent expression levels, potentially useful for downstream tasks such as disease classification.
\end{itemize}

\section{Method}

\begin{figure}[t]
    \centering
    \includegraphics[width=\linewidth]{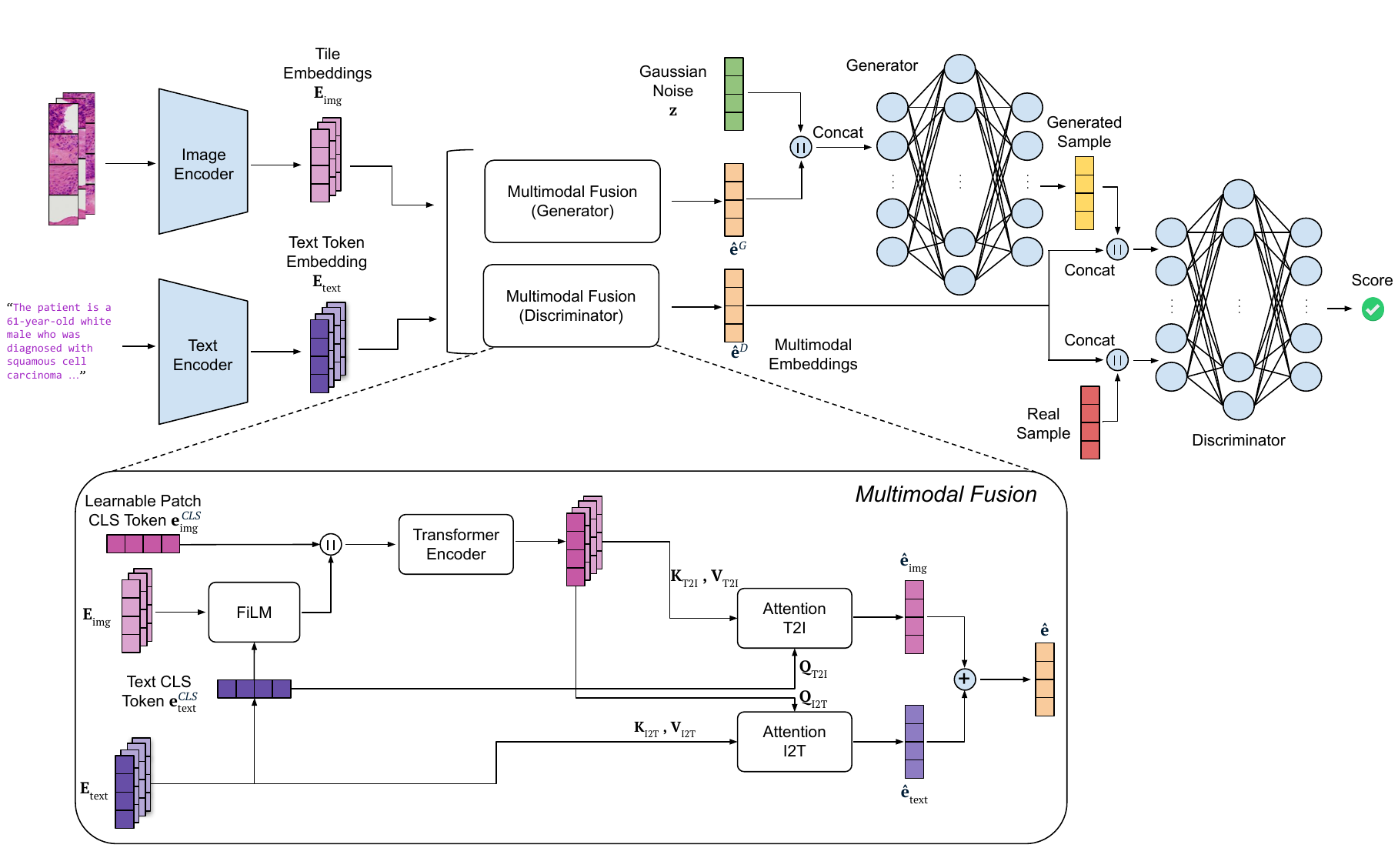}
    \caption{The architecture of our generative model. The input patches and textual descriptions are embedded with the Image and Text Encoder, respectively. These embeddings are combined in the Multimodal Fusion network to condition a WGAN-GP.}
    \label{fig:method}
\end{figure}

At a high level, \methodName{} consists of three main steps, as depicted in \cref{fig:method}. We first extract patches from tissue slides, embed each patch with an \emph{Image Encoder} and embed the clinical description with a \emph{Text Encoder}. These embeddings are then fused in a \emph{Multimodal Fusion} network that produces as output a single multimodal embedding capturing information from both modalities. Finally, the multimodal embedding is used as a conditioning vector for a Wasserstein GAN with Gradient Penalty (WGAN-GP)~\cite{gulrajani2017}, which generates gene expression profiles. All these steps are jointly trained in an end-to-end manner.

\subsection{Single-Modal Networks}

Our generative model is conditioned on the WSI patches $\{\patch^{(i)}\}$ and the textual description $\textualDescription$. To enable multimodal conditioning, we project each input modality into a latent space having the same dimension $\embeddingDim$. For the textual modality, we tokenize and embed the input using the \emph{Text Encoder}, that outputs an embedding matrix $\textTokenEmbeddings \in \mathbb{R}^{\numTokens \times \embeddingDim}$, where $\numTokens$ is the maximum number of tokens allowed for the textual description. The first row of this matrix is the Text CLS token $\textCLS \in \mathbb{R}^{\embeddingDim}$, which is directly modelled by the \emph{Text Encoder}. For the imaging modality, we randomly sample a subset $\{\patch^{(i_j)}\}_{j=0}^\numPatches$ of $\numPatches$ patches, which are processed by the \emph{Image Encoder}, producing as output an embedding matrix $\patchEmbeddings \in \mathbb{R}^{\numPatches \times \embeddingDim}$. As the \emph{Image Encoder} process each patch separately, there is no natural CLS token inside $\patchEmbeddings$. Thus, we use a learnable vector $\patchCLS \in \mathbb{R}^{\embeddingDim}$ as a surrogated Patch CLS token for the image modality.

\subsection{Multimodal Fusion}

Our Multimodal Fusion network takes as input the patch embeddings $\patchEmbeddings$ and the text tokens $\textTokenEmbeddings$, producing as output a joint representation $\multimodalEmbedding{} \in \mathbb{R}^{\embeddingDim}$:
\begin{equation}
    \multimodalEmbedding{} = MultimodalFusion(\patchEmbeddings, \textTokenEmbeddings).
\end{equation}

We first modulate the patch embeddings exploiting Feature-wise Linear Modulation (FiLM) \cite{perez2018film}, leveraging the Text CLS Token to condition the image features. Specifically, FiLM learns two functions $\gamma: \mathbb{R}^\embeddingDim \rightarrow \mathbb{R}^\embeddingDim$ and $\beta: \mathbb{R}^\embeddingDim \rightarrow \mathbb{R}^\embeddingDim$ that are used to compute a feature-wise affine transformation of each patch embedding:
\begin{equation}
    FiLM(\patchEmbeddings, \textCLS) = \gamma(\textCLS) \odot \patchEmbeddings_{i,c} + \beta(\textCLS).
\end{equation}
The result is a set of modulated patch embeddings, denoted by $\modulatedPatchEmbeddings \in \mathbb{R}^{\numPatches \times \embeddingDim}$, which emphasized features that are more relevant, according to the textual description. We then prepend the learnable Patch CLS Token $\patchCLS$ to the modulated patch embeddings and input these to a Transformer Encoder \cite{vaswani2017attention}. This enables us to capture relationships between different patches of the WSI and, at the same time, summarize the most relevant patterns of the WSI into the updated Patch CLS Token. Indeed, WSIs might contain patches that refer to the diagnosed disease and other patches that are not relevant. We denote with $\transformerPatchEmbeddings \in \mathbb{R}^{(\numPatches+1) \times \embeddingDim}$ the output of the Transformer Encoder, having as first row the updated Patch CLS Token, denoted by $\trasformerPatchCLS$.

Finally, to capture relationships between the two input modalities, we use a bidirectional Cross-Attention mechanism exploiting two Multi-Head Attention:
\begin{equation}
    \multimodalEmbedding{}_{img} = \mathit{MultiHeadAttention}_{T2I}(Q=\textCLS, K=\transformerPatchEmbeddings, V=\transformerPatchEmbeddings),
\end{equation}
\begin{equation}
    \multimodalEmbedding{}_{text} = \mathit{MultiHeadAttention}_{I2T}(Q=\trasformerPatchCLS, K=\textTokenEmbeddings, V=\textTokenEmbeddings),
\end{equation}
where $T2I$ and $I2T$ stands for $Text2Image$ and $Image2Text$, respectively. The Text CLS Token attends over the visual tokens to obtain $\multimodalEmbedding{}_{img}$, while the updated Patch CLS Token attends over the textual embeddings to produce $\multimodalEmbedding{}_{text}$. The final multimodal embedding is given by the sum of both: $\multimodalEmbedding{} = \multimodalEmbedding{}_{text} + \multimodalEmbedding{}_{img}$.

\subsection{Generative Model}
As generative model, we adopt the WGAN-GP architecture \cite{gulrajani2017}, which remains the state-of-the-art for transcriptomic data generation due to its ability to handle high-dimensional inputs with stable training dynamics. While diffusion-based models have recently begun to emerge in this domain \cite{lacan2024silico}, they are still in early exploratory stages and not yet widely adopted. Our choice of WGAN-GP is further supported by empirical comparisons, where alternative models such as VAEs are consistently underperformed in terms of sample quality and distributional fidelity.

We condition the WGAN-GP on the output of the Multimodal Fusion network. To obtain the two conditioning vectors (one for the generator and one for the discriminator) to be concatenated with both real and fake gene expression profiles, we use the same Multimodal Fusion network, with different learnable parameters $\theta_G$ and $\theta_D$:
\begin{equation}
    \multimodalEmbedding{G} = \mathit{MultimodalFusion}^{G}(\patchEmbeddings, \textTokenEmbeddings; \theta_G),
\end{equation}
\vspace{-4mm}
\begin{equation}
    \multimodalEmbedding{D} = \mathit{MultimodalFusion}^{D}(\patchEmbeddings, \textTokenEmbeddings; \theta_D),
\end{equation}

The Generator, a Multi-Layer Perceptron (MLP), takes a noise vector $\noise \sim \normalDistribution{\embeddingDim}$ and a multimodal embedding $\multimodalEmbedding{G}$ to produce  a gene expression profile $\geneExpressionOut{gen}$. The Discriminator, also an MLP, receives $\multimodalEmbedding{D}$ along with real and generated profiles, and learns to distinguish between them. This conditional setup guides the generator to produce biologically meaningful outputs semantically consistent with the input tissue slides and textual descriptions.

\section{Implementation Details}


\subsection{Data}

The input data consist of paired histopathology whole-slide images (WSIs), clinical metadata, and matched gene expression profiles, all retrieved from TCGA public repository\footnote{\url{https://www.cancer.gov/ccg/research/genome-sequencing/tcga}} and focused on nineteen different tumor types. Gene expression profiles, obtained from RNA sequencing, are quantified using FPKM (Fragments Per Kilobase of transcript per Million mapped reads), providing normalized measures of gene activity specific to the disease context. WSIs are ultra-high-resolution images of tissue sections, which are computationally infeasible to analyze in full, and are therefore typically subdivided into smaller tiles for downstream processing (\cref{fig:preprocessing}). Clinical metadata is available in JSON format and includes patient-specific information such as demographics, cancer subtype, and treatment history, all related to the disease condition. We convert this structured metadata into concise case summaries using a quantized instruction-tuned language model. We used a version of Llama3-8B fine-tuned on medical data ~\footnote{\url{https://huggingface.co/ContactDoctor/Bio-Medical-Llama-3-8B}}. Irrelevant fields are removed, and the remaining data is serialized into prompts, resulting in ~200-word descriptions that capture disease site, demographics, and experimental conditions. This step is essential to align the TCGA input format with the intended use case, where users may input a WSI and write a text description to obtain a plausible gene expression.

\subsection{Preprocessing}
\label{sec:preprocessing}

\begin{figure}[t]
    \centering
    \includegraphics[width=\linewidth]{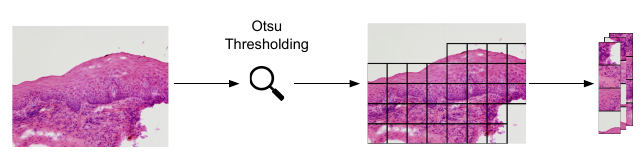}
    \caption{Preprocessing of the WSIs with Otsu Thresholding to extract valid image patches that contain at least 20\% of tissue content} 
    \label{fig:preprocessing}
\end{figure}

To prepare data for multimodal generation (\cref{fig:preprocessing}), we segment tissue regions from pathology slides using Otsu Thresholding \cite{otsu1975threshold} and extract high-resolution tiles of size $\patchSize = 256$, keeping only those tiles with more than 20\% of tissue content.
For the gene expression data, we remove genes with more than 90\% missing values and apply standard score normalization (z-score). 

\subsection{Embedding models and Hyperparameters}

We use UNI \cite{chen2024uni} as an Image Encoder, which offers a pretrained Vision Transformer for histopathology tiles. As a Text Encoder, we exploit Clinical ModernBERT \cite{lee2025clinical} \footnote{\url{https://huggingface.co/Simonlee711/Clinical_ModernBERT}} \footnote{Both UNI and Clinical ModernBERT are accessible through Huggingface}. As these two models have a different embedding size, \emph{i.e.} 1024 for UNI and 768 for Clinical ModernBERT, we add a linear projection layer to the outputs of the two models to align the embedding sizes to $\embeddingDim = 256$. In practice, as these models are already pretrained on a high amount of data, we train only the linear projection layer and freeze the pretrained layers. For the Generator and Discriminator, we use two MLPs having two hidden layers of size 256. To train our model, we select $\numPatches = 256$ random patches at each training step.

\section{Experiments}


We test our method on the TCGA dataset, that provides both tissue slides and clinical descriptions of each patient, associated with a gene expression profile derived from RNA-seq. For storage and computational purposes, we select only a subset of TCGA by keeping only the samples having a tissue slide of dimension less than 100 MB, for a total of 1944 clinical cases. 
For each patient, following the preprocessing steps described in \cref{sec:preprocessing}, we obtain expression values for 18,868 genes, where each gene corresponds to a column in the tabular matrix to be generated.
\subsection{Evaluation metrics}

We evaluate our approach with three classes of metrics: (i) \emph{unsupervised metrics}, (ii) \emph{detectability} and (iii) \emph{utility}.

\subsubsection{Unsupervised Metrics} 
\paragraph{Precision and Recall}\cite{prec_recall} are metrics used to assess the quality and relevance of the generated data with respect to the real data. Given the real dataset $\textbf{X} =\{ x_1, \ldots, x_n\}$ and the generated dataset $\hat{\textbf{X}} = \{\hat{x}_1, \ldots, \hat{x}_m \}$, the first step is to compute the manifolds for the real and generated samples $H^t_\textbf{X}$ and $H^t_{\hat{\textbf{X}}}$, respectively. This is done by forming a sphere around each point of the two datasets with a radius corresponding to the euclidean distance to \textit{t}-th closest neighbor within the same dataset. A binary function $f(h, H)$ is then defined to evaluate whether the sample $h$ falls within the manifold $H$.
Precision and recall can be computed as follows:

{\small
\begin{equation}
\text{precision}_t(\textbf{X}, \hat{\textbf{X}}) = \frac{1}{|\hat{\textbf{X}}|} \sum_{\hat{\textbf{x}} \in \hat{\textbf{X}}} f(\hat{\textbf{x}}, H^t_\textbf{X}),
\end{equation}
\begin{equation}
\text{recall}_t(\textbf{X}, \hat{\textbf{X}}) = \frac{1}{|\textbf{X}|} \sum_{\textbf{x} \in \textbf{X}} f(\textbf{x}, H^t_{\hat{\textbf{X}}}).
\end{equation}
}
We compute precision and recall considering the 10-th nearest neighbors. 

\paragraph{Structural Coherence} 

To assess structural quality, we compute the mean squared error (C. MSE) between real and synthetic gene-gene correlation matrices. Lower C. MSE values indicate better preservation of biologically meaningful co-expression patterns.



\subsubsection{Supervised Metrics} 
\paragraph{Detectability} The detectability measures how well a classfication model can distinguish generated from real samples. In our experiments, we use Logistic Regression and Multi-Layer Perceptron as classifiers and Accuracy and F1-Score as figures of merit. In particular, low values for Accuracy and F1-Score indicate high similarity between real and generated samples.

\paragraph{Utility} The utility measures the performance of models trained on downstream tasks. In our experiments, we train classification models on generated data to predict the disease type or the primary site given a gene expression profile and test these models on real data. Higher performance (accuracy and F1-Score) indicates a more realistic data generation.

\subsection{Competitors}
We compare our results with three different benchmarks: a Vanilla WGAN-GP, that is a WGAN-GP without any conditioning, a Conditional WGAN-GP, conditioned on two categorical variables, namely the disease type and the primary site, and a Conditional Variational Autoencoder (CVAE), conditioned on the same variables of the Conditional WGAN-GP. Please note that we cannot perform an utility evaluation on the Vanilla WGAN-GP as this is unconditional. Differently, by conditioning the other baselines on the disease type and the primary site, we can use the conditional variables as labels of the generated data to train the classification models.

\subsection{Experimental Setup}
We performed our experiments on a machine with the AMD Ryzen 1950X CPU, with 128 GB of RAM, using two NVIDIA A6000 GPUs with 48 GB of VRAM. We employed PyTorch \footnote{{\url{https://pytorch.org/}}} version 2.6.0 to define our model and perform the experimental evaluation, and CUDA version 12.4 for model training on the GPU devices. We perform an 80/20 train-test split, ensuring that the histopathology images, clinical metadata, and corresponding gene expression profiles are divided consistently across sets. We train all the models using a latent dimension $d = 256$ for the noise and with a batch size of 64.

\subsection{Main Results}
As shown in \cref{table:main-results}, our method demonstrates superior recall performance (0.8190) while maintaining balanced precision-recall characteristics. Although precision (0.7103) is lower than conditional WGAN-GP (0.8436), the higher recall indicates successful capture of dataset variability, effectively avoiding mode collapse common in GAN-based approaches \cite{kossale2022mode}.
Most notably, our method achieves exceptional gene correlation preservation with MSE of 0.0007, representing 98.2\% of improvement over the Conditional WGAN-GP (0.0038). 
This dramatic improvement demonstrates our model's ability to preserve biologically meaningful gene relationships.

Detectability analysis reveals that our synthetic data is significantly harder to distinguish from real data using LR, indicating high-quality synthesis despite genomic data complexity. However, sophisticated classifiers like MLPs maintain high detectability, presenting an area for future investigation.

Our method consistently outperforms baselines across all utility metrics. For disease type classification, we achieve 16.9\% improvement in RF accuracy and 11.1\% improvement in F1 score. Primary site classification shows even greater improvements: 13.4\% in RF accuracy and 10.0\% in F1 score. These improvements stem from our model's powerful embedding-based conditioning, which generates semantically meaningful synthetic data that maintains relevant biological patterns for downstream applications.

\begin{table}[t]
    \centering
    \caption{Results of our models on the TCGA dataset, compared with three different benchmarks: a WGAN-GP (\emph{Vanilla WGAN-GP}), a WGAN-GP (\emph{Cond. WGAN-GP}) and CVAE, both conditioned on disease type and primary site. Mean and standard deviation across 10 generation runs are reported. Best results are highlighted in bold.
    } 
    \begin{tabular}{>{\centering\arraybackslash}p{2.3cm}|>{\centering\arraybackslash}p{2.3cm}|>{\centering\arraybackslash}p{2.3cm}|>{\centering\arraybackslash}p{2.3cm}|>{\centering\arraybackslash}p{2.3cm}}
         Metric & Vanilla WGAN-GP & Cond. WGAN-GP & CVAE & Proposed  \\
         \hline
         \multicolumn{5}{c}{Unsupervised Metrics} \\
         \hline
         Precision $\uparrow$ & 0.8282(0.0000) & \textbf{0.8436(0.0051)} & 0.4892(0.0015) & 0.7103(0.0047) \\
         Recall $\uparrow$ & 0.6477(0.0031) & 0.7372(0.0038) & 0.7956(0.0131) & \textbf{0.8190(0.0068)} \\
         C. MSE $\downarrow$ & 0.0095(0.0012) & 0.0038(0.0001) & 0.2837(0.0027) & \textbf{0.0007(0.0000)} \\
         \hline
         \multicolumn{5}{c}{Detectability} \\
         \hline
         LR (Acc.) $\downarrow$ & 0.8459(0.0142) & 0.8801(0.0014) & 0.8931(0.0077) & \textbf{0.5236(0.0179)} \\
         LR (F1) $\downarrow$ & 0.8662(0.0107) & 0.8925(0.0119) & 0.9003(0.0070) & \textbf{0.5444(0.0286)} \\
         MLP (Acc.) $\downarrow$ & 
         0.9887(0.0040)
         & 0.9904(0.0019) & 0.9992(0.0006) & \textbf{0.9797(0.0141)} \\
         MLP (F1) $\downarrow$ & 0.9886(0.0040) & 0.9905(0.0019) & 0.9992(0.0006) & \textbf{0.9794(0.0145)}\\
         \hline
         \multicolumn{5}{c}{Utility - Disease Type} \\
         \hline
         RF (Acc.) $\uparrow$ & - & 0.6367(0.0403) & 0.5725(0.0196) & \textbf{0.7449(0.0127)} \\
         RF (F1) $\uparrow$ & - & 0.7422(0.0287) & 0.6698(0.0165) & \textbf{0.8249(0.0087)} \\
         MLP (Acc.) $\uparrow$ & - & 0.9262(0.0048) & 0.9031(0.0007) & \textbf{0.9354(0.0072)} \\
         MLP (F1) $\uparrow$ & - & 0.9247(0.0053) & 0.8803(0.0006) & \textbf{0.9322(0.0072)} \\
         \hline
         \multicolumn{5}{c}{Utility - Primary Site} \\
        \hline
         RF (Acc.) $\uparrow$ & - & 0.6118(0.0244) & 0.4228(0.0075) & \textbf{0.6936(0.0061)} \\
         RF (F1) $\uparrow$ & - & 0.6800(0.0201) & 0.4726(0.0049) & \textbf{0.7478(0.0090)} \\
         MLP (Acc.) $\uparrow$ & - & 0.8254(0.0091) & 0.6782(0.0176) & \textbf{0.8436(0.0069)} \\
         MLP (F1) $\uparrow$ & - & 0.8094(0.0070) & 0.5735(0.0222) & \textbf{0.8264(0.0072)} \\
         \hline
    \end{tabular}
    \label{table:main-results}
\end{table}


\subsection{Ablation Study}
To evaluate the contribution of both text and image modalities to the generative process, we conducted an ablation study comparing different conditioning strategies. In particular, we investigated whether our proposed image encoding strategy provides a more informative signal than simpler alternatives, such as the mean of the patch embeddings or the use of a CLS token for the input text. The results in \cref{table:ablation-study} highlight that image-based conditioning plays a crucial role in balancing precision and recall, as demonstrated by the significantly more stable values across all image-informed variants. In contrast, conditioning solely on text yields a high precision (0.9513) but markedly lower recall (0.5205), suggesting that it leads to more selective but less comprehensive generations. From a utility perspective, image conditioning consistently yields higher accuracy and F1 scores, indicating that it carries richer semantic information for downstream tasks. Our full model achieves the highest overall performance in both utility prediction and detectability metrics, confirming its robustness and practical utility. While we acknowledge the trade-offs between unsupervised and supervised metrics, our approach proves to be the most effective in producing realistic and functionally valuable samples.
\begin{table}[t]
\centering
\caption{Ablation study. We compare our conditional generative models with different strategies for the conditioning vector of the WGAN-GP: (i) the mean of the patch embeddings (\emph{Mean Image}), (ii) the CLS token embedding of the input text (\emph{Text CLS Token}), (iii) a Transformer Encoder for the patch embeddings (\emph{Patch Transformer}), (iv) our model without the final Cross Attention (\emph{FiLM}), (v) our model without FiLM (\emph{Cross Attention}). Results are the mean of 10 generation runs.}
\resizebox{\textwidth}{!}{%
\begin{tabular}{|>{\centering\arraybackslash}p{2.8cm}|>{\centering\arraybackslash}p{1.3cm}|>{\centering\arraybackslash}p{1.3cm}|>{\centering\arraybackslash}p{1.3cm}|>{\centering\arraybackslash}p{1.3cm}|>{\centering\arraybackslash}p{1.3cm}|>{\centering\arraybackslash}p{1.3cm}|>{\centering\arraybackslash}p{1.3cm}|>{\centering\arraybackslash}p{1.3cm}|}
\hline
\multirow{2}{*}{Method} & \multicolumn{3}{c|}{Unsupervised} & \multicolumn{2}{c|}{Detectability (LR)} & \multicolumn{2}{c|}{Utility (RF)}\\
\cline{2-8}
& Prec. & Recall & C. MSE & Acc. & F1 & Acc. & F1 \\
\hline
Mean Image & 0.8128 & 0.7026 & 0.0014 & 0.5449 & 0.5730 & 0.7167 & 0.8011 \\
\hline
Text CLS Token & \textbf{0.9513} & 0.5205 & 0.0045 & 0.5571 & 0.5797 & 0.5000 & 0.6330 \\
\hline
Patch Transformer & 0.8500 & 0.7821 & 0.0013 & 0.5692 & 0.5772 & 0.7333 & 0.8158 \\
\hline
FiLM & 0.7910 & \textbf{0.8821} & 0.0015 & 0.5853 & 0.5967 & 0.6513 & 0.7555 \\
\hline
Cross Attention & 0.7000 & 0.8346 & 0.0011 & 0.5526 & 0.5524 & 0.7256 & 0.8112 \\
\hhline{|=|=|=|=|=|=|=|=|}
Ours & 0.7103 & 0.8190 & \textbf{0.0007} & \textbf{0.5236} & \textbf{0.5444} & \textbf{0.7449} & \textbf{0.8249} \\
\hline
\end{tabular}
}
\label{table:ablation-study}
\end{table}


\section{Conclusions}

This work introduced \methodName{}, the first multimodal generative framework that conditions gene expression synthesis on both histopathology images and clinical metadata. The model achieves strong performance across standard evaluation metrics, producing statistically realistic and biologically coherent gene expression profiles.

Through a detailed ablation study, we quantified the contribution of each modality to the generative process, revealing the crucial role of histopathology images. These findings demonstrate that tissue morphology encodes rich semantic information, which significantly enhances the biological fidelity of the generated profiles. This analysis also improves the interpretability of our model by clarifying how visual and textual inputs influence the generation pipeline.

Future work will extend this framework to generate histopathology images from gene expression profiles and clinical data. This bidirectional capability would enable researchers to simulate how genetic changes appear in tissue samples, advancing our understanding of disease mechanisms and supporting the development of cross-modal AI tools for precision medicine.

\bibliographystyle{splncs04}
\bibliography{main}
\end{document}